\pdfoutput=1

\documentclass[11pt]{article}

\usepackage[preprint]{acl}

\usepackage{times}
\usepackage{latexsym}

\usepackage[T1]{fontenc}

\usepackage[utf8]{inputenc}

\usepackage{microtype}

\usepackage{inconsolata}

\usepackage{graphicx}
\usepackage{multirow} 
\usepackage{inconsolata}
\usepackage{xspace}
\newcommand{\modelname}[0]{AnyTrans\xspace}
\newcommand{\taskname}[0]{{\textsc{Tati}}\xspace}

\title{\modelname: Translate AnyText in the Image with Large Scale Models}

\author{{Zhipeng Qian$^1$\footnotemark[1],\footnotemark[3]} , {\bf Pei Zhang$^{2,3}$\footnotemark[1]} , {Baosong Yang$^{2}$} , {Kai Fan$^{2}$}, {Yiwei Ma$^{1}$} , \\
 {\bf Derek F. Wong$^{3}$} , {\bf  Xiaoshuai Sun $^1$\footnotemark[2]} , {\bf Rongrong Ji$^1$},\\
        $^1$Key Laboratory of Multimedia Trusted Perception and Efficient Computing, \\Ministry of Education of China, Xiamen University, \\$^2$ Alibaba Group Inc, $^3$ NLP2CT Lab, University of Macau \\
        }
 
\usepackage{adjustbox}
\begin{document}
\maketitle
\renewcommand{\thefootnote}{\fnsymbol{footnote}} 
\footnotetext[1]{These authors contributed equally.}
\footnotetext[2]{The corresponding author.}
\footnotetext[3]{Work done during internship at Institute for Intelligent Computing, Alibaba Group Inc.}

\begin{abstract}
This paper introduces \modelname, an all-encompassing framework for the task--Translate AnyText in the Image (\taskname), which includes multilingual text translation and text fusion within images. 
Our framework leverages the strengths of large-scale models, such as Large Language Models (LLMs) and text-guided diffusion models, to incorporate contextual cues from both textual and visual elements during translation. 
The few-shot learning capability of LLMs allows for the translation of fragmented texts by considering the overall context. 
Meanwhile, the advanced inpainting and editing abilities of diffusion models make it possible to fuse translated text seamlessly into the original image while preserving its style and realism. 
Additionally, our framework can be constructed entirely using open-source models and requires no training, making it highly accessible and easily expandable. 
To encourage advancement in the \taskname task, we have meticulously compiled a test dataset called MTIT6, which consists of multilingual text image translation data from six language pairs.


\end{abstract}

\section{Introduction}

Translating AnyText in the Image (\taskname) has become an essential tool in our daily lives, transforming how we interact with the world. 
This capability extends to a wide range of applications, from facilitating cross-cultural communication to supporting education, and playing a significant role in global business operations. 
Falling under the umbrella of multi-modal machine translation (MMT)~\cite{Elliott_Frank_Sima’an_Specia_2016,Calixto_Li_Campbell_2017,Elliott_Kádár_2017,Libovický_Helcl_Mareček_2018,Sulubacak_Caglayan_Grönroos_Rouhe_Elliott_Specia_Tiedemann_2019}, the process of translating text in images is commonly known as Text Image Translation (TIT)~\cite{Ma_Zhang_Tu_Han_Wu_Zhao_Zhou_2022,Lan_Yu_Li_Zhang_Luan_Wang_Huang_Su_2023}. 
TIT aims to accurately convert text in source images into desired target languages. 

However, we argue that translated text alone is insufficient. 
A seamless integration of text and image is crucial for effectively conveying the intended message. 
Thus, we believe that our proposed task, Translate AnyText in the Image (\taskname), better aligns with practical needs. 
It not only aims to translate textual content within an image but also maintains the visual coherence and intrinsic harmony of text and graphic elements, thereby enhancing the overall comprehensibility of texts in images.

\begin{figure}[t]
    \centering
    \includegraphics[width=\columnwidth]{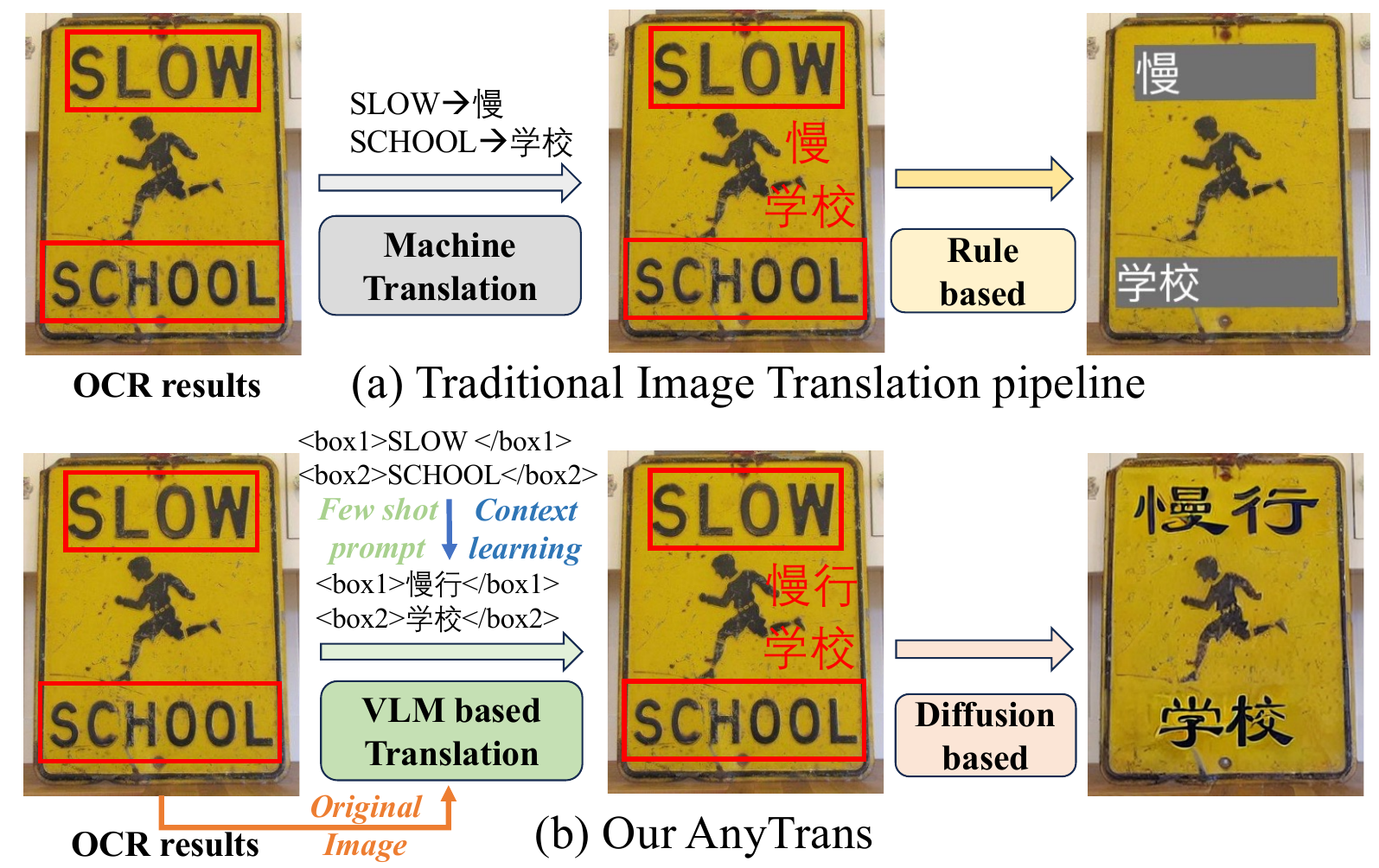}
    \vspace{-0.5cm}
    \caption{
Comparison between traditional image translation pipeline and our \modelname. 
Our \modelname combines image information and context for more accurate translation and generates more realistic text.}
\vspace{-0.5cm}
    \label{fig:intro2}
\end{figure}

Current popular products, such as Google Image Translation\footnote{\scriptsize{\url{https://translate.google.com}}}, Microsoft Image Translation\footnote{\scriptsize{\url{https://www.microsoft.com/en-us/translator/apps/}}}, and Apple iOS Image Translation\footnote{\scriptsize{\url{https://support.apple.com/zh-cn/guide/iphone/iphea8b95631/ios}}}, have made significant progress in translating text within images. 
However, as illustrated in Figure~\ref{fig:intro2} (a), Microsoft Image Translation, for instance, utilizes traditional machine translation to translate text recognized by OCR models. 
It then employs a simple rule to insert the translated text back into the original image. 
Unfortunately, this approach often overlooks the contextual relationship between textual elements within images. 
This oversight can result in inaccurate translations and visual inconsistencies, thereby compromising the authenticity of the newly generated image.

To address the identified shortcomings, our framework illustrated in Figure~\ref{fig:intro2} (b) significantly diverges from conventional text translation tasks in images. 
By leveraging the advanced contextual comprehension capabilities of LLMs, our approach achieves superior translation accuracy. 
Alternatively, the integration of a vision language model (VLM) may allow a dual consideration of both visual and textual contexts within the source images, further enhancing translation quality. 

Our methodology unfolds in three consecutive steps. 
Initially, we utilize the latest PP-OCR~\cite{du2020pp} to accurately locate the text within the image and decipher its content. 
This step is crucial for determining the exact area for text editing and translating the text content precisely.
Secondly, once the text is identified, we employ a few-shot prompt learning strategy that enables (visual) language models to maintain the format during contextual translation. 
This approach ensures that the translation is both contextually appropriate and linguistically accurate. 
Finally, we apply a modified AnyText~\cite{tuo2023anytext} to render the translated text back into the original image. 
In this phase, the translated text is fused into its original location, identified during the initial step. 
We propose resizing the anticipated text box by considering the length of the detected box, the original source text, and the translated target text. 
This modification maximizes the preservation of the original image's style and produces a clean, new image. 
As shown in Figure~\ref{fig:intro2} (b), our method does achieve superior translation quality and visual effects while preserving the image's legibility and aesthetic appeal. 
The new text seamlessly blends with the original visual context, maintaining both coherence and style. 

Our main contributions are as follows:

\paragraph{(1)} We present an integrated framework for the task--Translate AnyText in the Image (\taskname), consisting of three key steps: source text detection and recognition, text image translation, and target text fusion. 

\paragraph{(2)} Our method is training-free and can be built entirely on open-source models, yet it delivers results that are comparable to or even surpass those of commercial, proprietary products.

\paragraph{(3)} We constructed a multilingual text image translation test dataset called MTIT6, which consists of translation data in six language pairs and is manually sequenced by humans, promoting the field of image translation.


\section{Related Works}

\subsection{Text Image Translation and Multilingual Translation}

The field of multimodal machine translation (MMT) \cite{Caglayan_Barrault_Bougares_2016,Huang_Liu_Shiang_Oh_Dyer_2016,Libovický_Helcl_2017,Calixto_Li_Campbell_2017,Su_Chen_Jiang_Zhou_Lin_Ge_Wu_Lai_2021} has witnessed remarkable advancements in recent years, catalyzing a surge in scholarly and industry interest. 
The prevailing practical demand for MMT is the translation of text within images, known as text image translation (TIT) \cite{Ma_Zhang_Tu_Han_Wu_Zhao_Zhou_2022,Mansimov_Stern_Chen_Firat_Uszkoreit_Jain_2020,Jain_Firat_Ge_Liang,Lan_Yu_Li_Zhang_Luan_Wang_Huang_Su_2023}. 
However, TIT leaves the image unchanged, while integrating text translations directly into images is essential for helping users understand the meaning of both text and visuals. 
Taking these factors into account, we believe that our proposed \taskname task is more aligned with practical requirements. 

Meanwhile, Large Language Models (LLMs) \cite{Gao_He_Wu_Wang_2024,Vilar_Freitag_Cherry_Luo_Ratnakar_Foster_2022,Zeng_Meng_Yin_Zhou_2023, Wu_Cheng_Wang_Li_2021} have shown impressive multilingual translation proficiency. 
Integrating multilingual translation \cite{Dong_Wu_He_Yu_Wang_2015,Firat_Cho_Bengio_2016,Neubig_Hu_2018,Chen_Liu_Cheng_Li_2017,Chen_Liu_Li_2022,Cheng_2019} with image-to-image translation opens vast opportunities and has wide-ranging applications, such as in cross-border e-commerce platforms, among others.

\begin{figure*}[t]
\centering 
\includegraphics[width=0.95\textwidth]{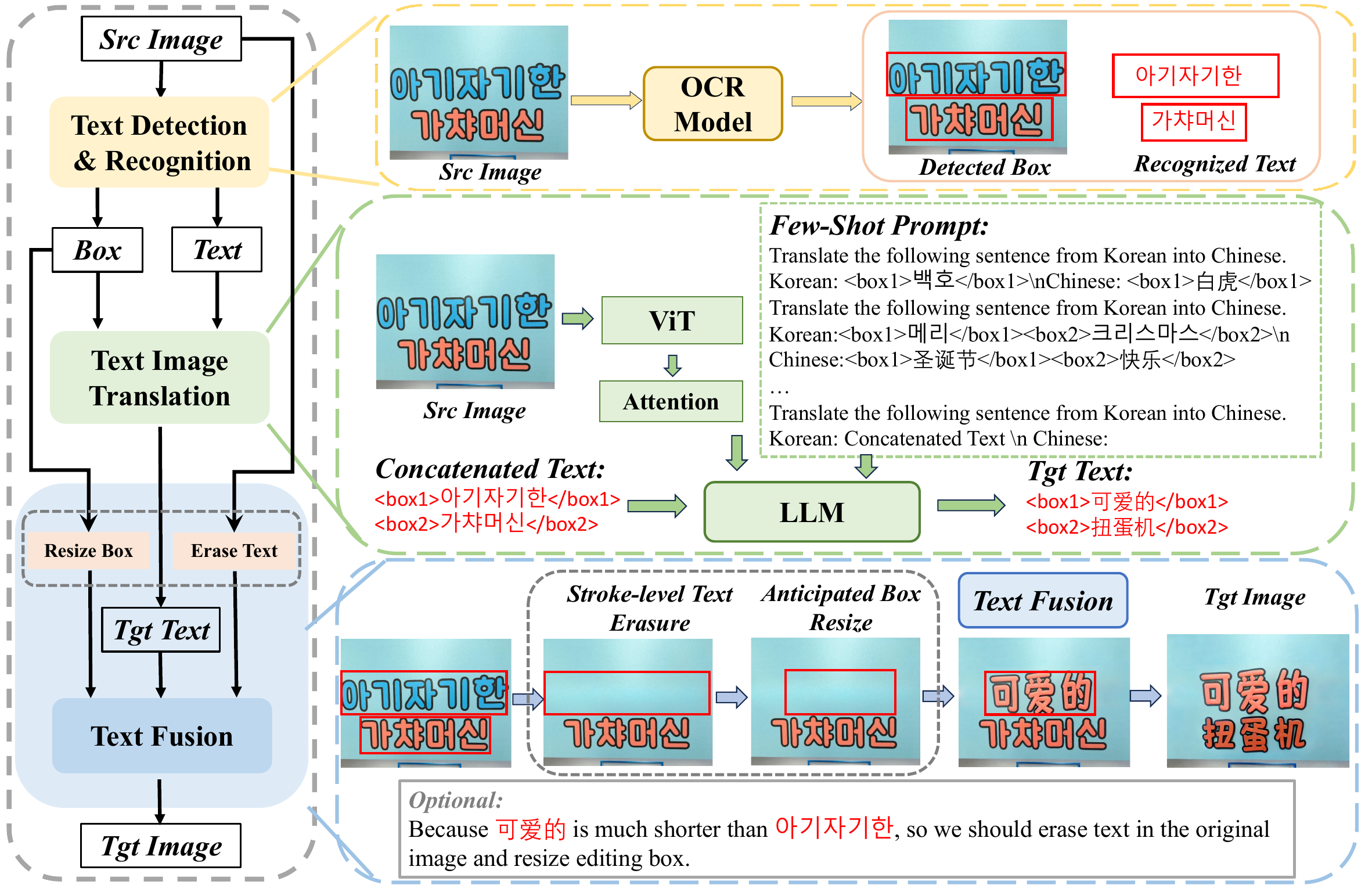}
\vspace{-0.3cm}
  \caption{An overview of \modelname. Our translation framework is built around three key components: firstly, Text Detection and Recognition utilizing an offline OCR model; secondly, Text Image Translation using (vision) LLMs; and finally, Text Fusion using the modified AnyText.}
  \vspace{-0.4cm}
  \label{fig:overview}
\end{figure*}

\subsection{Text Editing in Images}

Recent advancements in image processing have seen a burgeoning interest in text editing \cite{Yang_Liu_Yang_Guo_2018,Wu_Zhang_Liu_Han_Liu_Ding_Bai_2019,he2023wordart,Zhu_Wang_Shi_Zhang_Jiao_Tang,Ma_Zhao_Chen_Wang_Niu_Lu_Lin_2023, chen2024textdiffuser, Chen_Huang_Lv_Cui_Chen_Wei_2023,Couairon_Verbeek_Schwenk_Cord_2022,tuo2023anytext} within images.
Numerous methods leveraging Generative Adversarial Networks (GANs) have emerged for scene text editing, aiming to transform the text within a scene image to a specified target while retaining the authentic style. 
Despite their innovations, GAN-based approaches~\cite{Wu_Zhang_Liu_Han_Liu_Ding_Bai_2019,Goodfellow_Pouget-Abadie_Mirza_Xu_Warde-Farley_Ozair_Courville_Bengio_2017,Mirza_Osindero_2014,Zhu_Park_Isola_Efros_2017,Yang_Liu_Wang_Guo_2018, Azadi_Fisher_Kim_Wang_Shechtman_Darrell_2018} struggle to edit images featuring intricate scenes or a multitude of diverse elements. 
The recent development of diffusion models~\cite{Saharia_Chan_Chang_Lee_Ho_Salimans_Fleet_Norouzi_2022, Rombach_Blattmann_Lorenz_Esser_Ommer_2022,Chung_Sim_Ye_2022,Zhang_Ji_Zhang_Yu_Jaakkola_Chang_2023, Nichol_Dhariwal_Ramesh_Shyam_Mishkin_McGrew_Sutskever_Chen,Avrahami_Lischinski_Fried_2022,Yang_Gu_Zhang_Zhang_Chen_Sun_Chen_Wen_2022,zhang2023adding,Mou_Wang_Xie_Zhang_Qi_Shan_Qie_2023} allows for the generation of images of exceptional quality and diversity. 

Frameworks such as ControlNet \cite{zhang2023adding} and T2IAdapter \cite{Mou_Wang_Xie_Zhang_Qi_Shan_Qie_2023} have harnessed auxiliary cues like color maps, and segmentation maps to steer the image generation process, achieving remarkable levels of control and image quality. 
Galvanized by these advances, a series of text-centric image editing techniques~\cite{Zhu_Wang_Shi_Zhang_Jiao_Tang,Ma_Zhao_Chen_Wang_Niu_Lu_Lin_2023, chen2024textdiffuser, Chen_Huang_Lv_Cui_Chen_Wei_2023,Couairon_Verbeek_Schwenk_Cord_2022,tuo2023anytext} have been introduced based on diffusion models. 
Among these, AnyText \cite{tuo2023anytext} stands out for its proficient multilingual text editing capabilities, producing impressive results in text rendering and manipulation. 
The advancements of these technologies seamlessly enable the realization of \taskname task, facilitating a more intuitive and efficient process.

\section{Methodology}
In this section, we will detail each component of our \modelname. 
Following the module order shown in Figure~\ref{fig:overview}, we begin by introducing the detection and recognition of text in the image. 
Following this, we introduce how to leverage (vision) LLMs for translation. 
Lastly, we describe the text editing process informed by the translation outcomes.

\subsection{Text Detection and Recognition}

As illustrated in the \textit{Text Detection \& Recognition} section of Figure~\ref{fig:overview}, to accomplish our image-to-image translation task, we first need to detect the position of the text in the image and recognize its content. 
Essentially, this procedure involves text detection \cite{He_Liao_Yang_Zhong_Tang_Cheng_Yao_Wang_Bai_2021,Liao_Wan_Yao_Chen_Bai_2020,Lyu_Yao_Wu_Yan_Bai_2018,Ma_Shao_Ye_Wang_Wang_Zheng_Xue_2018,Zhou_Yao_Wen_Wang_Zhou_He_Liang_2017} and recognition \cite{Bautista_Atienza_2022,Li_Lv_Chen_Cui_Lu_Florencio_Cha_Li_Wei_2021,Shi_Bai_Yao_2017,yu2021benchmarking,Yu_Wang_Li_Xue_2023}, which embodies a classic OCR endeavour. 
Although VLM also has a certain degree of OCR capability, its capability lags far behind traditional OCR models~\cite{liu2023hidden}. 
So we harness the capabilities of the pre-trained OCR model, which excels in both text detection and recognition. 
Subsequently, the outcomes of OCR are fed into subsequent modules for translation and text editing.

\subsection{Beyond Box-level Text Translation}

\begin{figure}[t]
    \centering
    \includegraphics[width=\columnwidth]{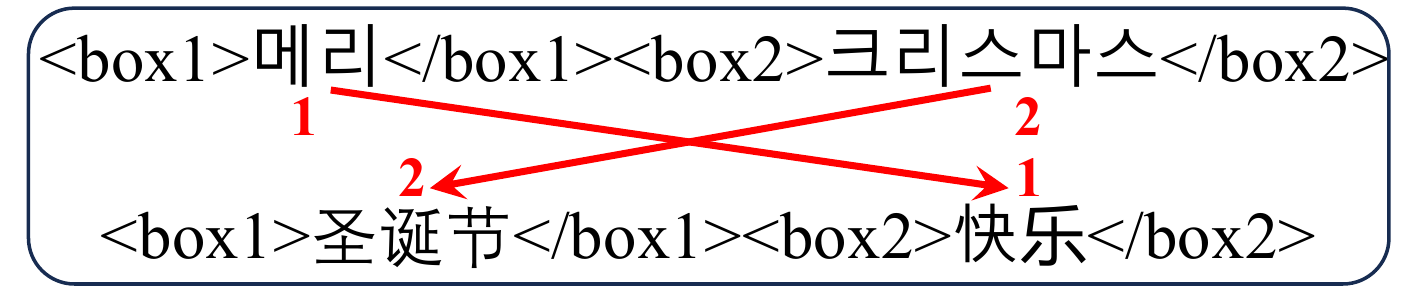}
    \vspace{-0.7cm}
    \caption{A prompt example from Korean to Chinese. In Chinese, the order of the two words should be switched.}
    \vspace{-0.7cm}
    \label{fig:prompt_example}
\end{figure}

Building on the recognition outcomes obtained from the OCR module, our next step involves translating the textual content into the desired target language. 
It is important to note that the OCR system processes and retrieves text content sequentially, which means the extracted sequence may not always reflect the true semantic order. 
This presents significant challenges for traditional translation models, which often struggle to accurately interpret the broader context and semantic connections between individual text segments. 
For instance, as illustrated in Figure~\ref{fig:intro2} (a), the word ``SLOW'' in an image should convey the meaning ``slow down for passing students''. 
However, traditional translation pipelines only translate the text within each isolated box, failing to grasp the context and leading to poor translations.

Fortunately, the landscape of translation has undergone a seismic shift with the emergence of large language models (LLMs), which exhibit a markedly enhanced ability to understand context and generate coherent translations. 
With their powerful multilingual and instruction-following capabilities, LLMs can be seamlessly integrated into our multilingual image translation framework without additional training. 
By employing a few-shot prompt strategy, we can enable the translation of multiple text segments in a more coherent manner.

Therefore, we integrated the LLM into the core of our proposed framework. 
Particularly, as shown in Figure~\ref{fig:prompt_example}, for texts within an image identified by OCR, we concatenate them into a long text sequence using HTML-style tags \texttt{<box{idx}></box{idx}>} to retain the positional information of the detected text. 
The translated sentence should be organized in the same format, but with the word order adjusted accordingly. 
In practice, we use five-shot demonstrations for each language pair in the instruction prompt to help the LLM understand our designed format.

Additionally, while multiple translation options may exist for a given text, the entire text sequences alone may not fully disambiguate the meaning. 
Therefore, incorporating visual information from images is also crucial. 
To address this, we have explored the supportive role of using a vision LLM in text translation. 
This method leverages the comprehensive visual information contained in images to refine the quality of the translation.

\begin{figure}[t]
    \centering
    \includegraphics[width=\columnwidth]{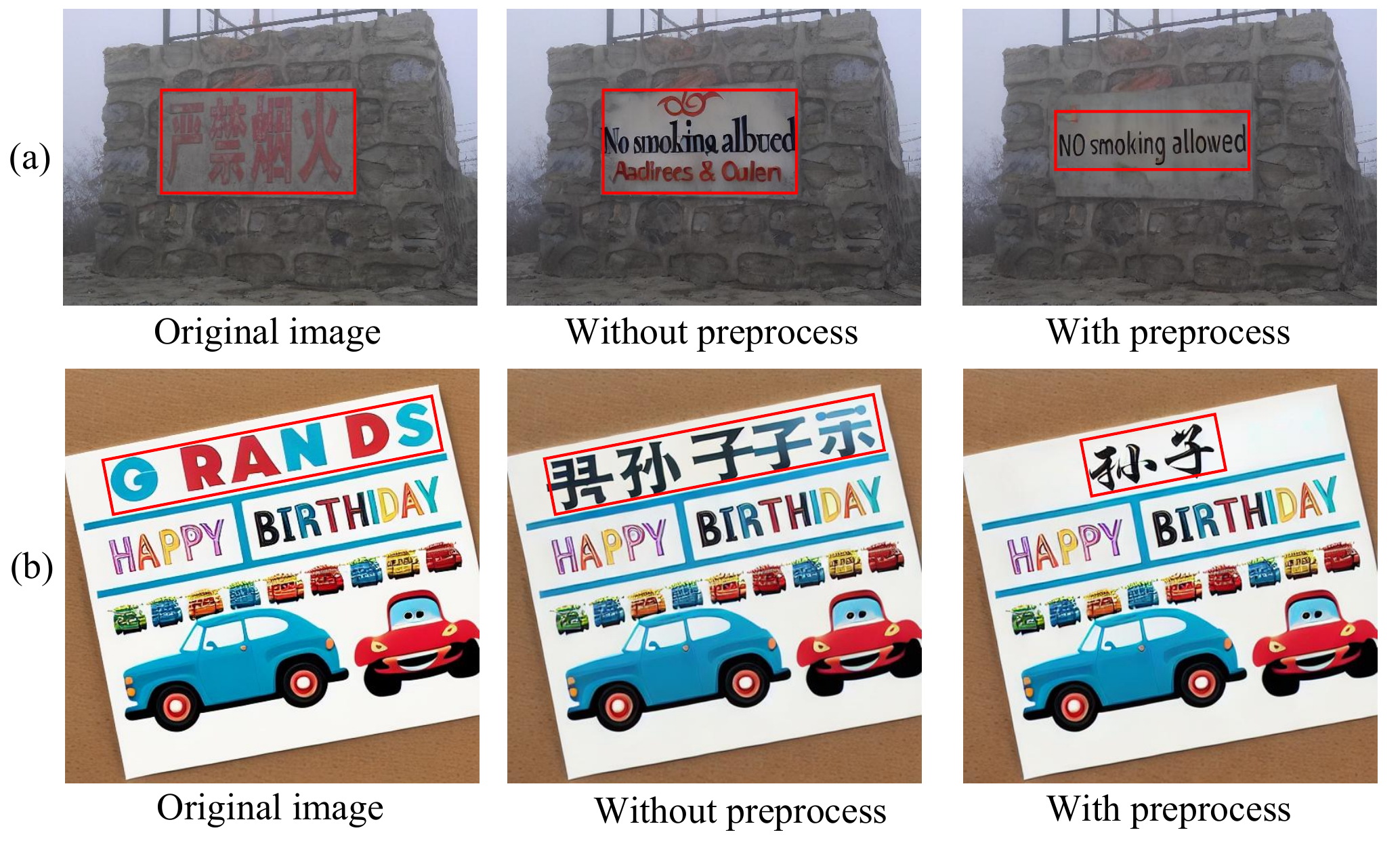}
    \vspace{-0.6cm}
    \caption{
Preprocessing for AnyText is crucial for producing accurate and authentic text, especially in scenarios where there is a significant disparity in text length before and after translation.}
\vspace{-0.6cm}
    \label{fig:long_shot}
\end{figure}

\subsection{Text Fusion in Image}
The final module in our framework involves generating a new image with the translated texts. 
To achieve a cohesive visual effect, we propose integrating the translated texts into the original image, placing them precisely where the original text appeared. 
This ensures that the translated text not only communicates the intended message but also harmonizes with the visual context of the image.

Traditional rule-based algorithms for fusing text into images exhibit several significant drawbacks, including compromising the integrity of the image background, limiting outputs to a singular font style, and resulting in a final appearance that often lacks realism. 
Instead, we adopt the technique of diffusion model, which enables natural text editing within images. 
Specifically, for our text editing process, we propose a multilingual text editing method built on Anytext \cite{tuo2023anytext}. 

In the original Anytext, the areas designated for editing are the detection boxes identified by OCR, and the input text is the translated sentence.
However, Anytext is particularly sensitive to the length of the input text designated for rendering. 
As shown in Figure~\ref{fig:long_shot}, the quality of the generated text is significantly impacted by the length ratio between the detected box and the input text. 
When this ratio deviates too far from 1, the vacant area tends to be filled with irrelevant content, significantly compromising both the visual effect and the translation quality.

\paragraph{Stroke-level Text Erasure} To address this issue, as illustrated in the \textit{Text Fusion} section of Figure~\ref{fig:overview}, we first apply stroke-level text erasure~\cite{9854177}. 
Unlike the end-to-end text editing approach used in Anytext, we decompose the process into two sub-steps. 
The first step involves applying a fine-grained inpainting method specifically designed to remove the strokes of characters or letters in the original texts. 
This method can successfully remove multi-line texts with minimal line spacing, resulting in a cleaner visual effect. 

\paragraph{Anticipated Box Resize} To address the length ratio issue and further avoid conflicts between adjacent lines, we propose an OCR box resizing preprocessing step for the anticipated target box. 
Specifically, if the word count ratio between the pre and post-translation text exceeds 1.2 or is less than 0.8, we will adjust the length or width of the anticipated box based on the ratio.
This process requires some customization depending on the language pair. 
For example, in zh-en translations, we assume the length of a Chinese character to be 2.5 times that of an English letter, given the fact that larger size for a single Chinese character. 
In the end, the fusion of target text is applied to the erased area.


\begin{table*}[t]
\resizebox{\textwidth}{!}{
\begin{tabular}{|c|ccc|ccc|ccc|}
\hline
\multirow{3}{*}{\textbf{Methods}}                                 & \multicolumn{3}{c|}{\textbf{\textbf{zh}$\rightarrow$\textbf{en}}}                                 & \multicolumn{3}{c|}{\textbf{zh}$\rightarrow$\textbf{ko}}                                & \multicolumn{3}{c|}{\textbf{zh}$\rightarrow$\textbf{ja}}                                 \\ \cline{2-10} 
                                                                  & \multicolumn{2}{c|}{\textbf{I2T}}                   & \textbf{I2I}  & \multicolumn{2}{c|}{\textbf{I2T}}                   & \textbf{I2I}  & \multicolumn{2}{c|}{\textbf{I2T}}                   & \textbf{I2I}  \\ \cline{2-10} 
                                                                  & \textbf{BLEU} & \multicolumn{1}{c|}{\textbf{COMET}} & \textbf{BLEU} & \textbf{BLEU} & \multicolumn{1}{c|}{\textbf{COMET}} & \textbf{BLEU} & \textbf{BLEU} & \multicolumn{1}{c|}{\textbf{COMET}} & \textbf{BLEU} \\ \hline
nllb-200(3.3B)                                                    & 29.7          & \multicolumn{1}{c|}{66.3}           & 22.2         & 20.1         & \multicolumn{1}{c|}{72.5}           & 11.4          & 24.9          & \multicolumn{1}{c|}{77.20}           & 13.1         \\ \hline
m2m100(1.2B)                                                      & 33.1         & \multicolumn{1}{c|}{66.1}           & 23.8         & 18.1          & \multicolumn{1}{c|}{71.1}           & 10.9          & 29.4          & \multicolumn{1}{c|}{79.60}           & 14.8          \\ \hline
mc-tit                                                     &41.6       & \multicolumn{1}{c|}{70.5}           &          &       & \multicolumn{1}{c|}{}           &          &         & \multicolumn{1}{c|}{}          &          \\ \hline
\begin{tabular}[c]{@{}c@{}}qwen1.5-7B-chat\end{tabular} & 37.4          & \multicolumn{1}{c|}{73.4}           & 26.5        & 11.4          & \multicolumn{1}{c|}{70.4}           & 5.5         & 31.2          & \multicolumn{1}{c|}{80.9}          & 20.6          \\ \hline
\begin{tabular}[c]{@{}c@{}}qwen1.5-14B-chat\end{tabular} & 38.8          & \multicolumn{1}{c|}{74.6}           & 28.0        & 16.1          & \multicolumn{1}{c|}{72.9}           & 8.3         & 30.7          & \multicolumn{1}{c|}{79.3}          & 19.8         \\ \hline
\begin{tabular}[c]{@{}c@{}}qwen1.5-110B-chat\end{tabular} & 43.8          & \multicolumn{1}{c|}{76.3}           & 30.6         & 17.1          & \multicolumn{1}{c|}{74.3}           & 9.3         & \textbf{35.4}          & \multicolumn{1}{c|}{\textbf{83.1}}           & \textbf{21.9}          \\ \hline
\begin{tabular}[c]{@{}c@{}}qwen-max\end{tabular}          & 44.0          & \multicolumn{1}{c|}{77.2}           &31.2       & 23.5          & \multicolumn{1}{c|}{75.1}           & 15.1          & 33.5          & \multicolumn{1}{c|}{81.3}           & 20.9          \\ \hline
\begin{tabular}[c]{@{}c@{}}qwen-vl-max\end{tabular}      & \textbf{48.7}          & \multicolumn{1}{c|}{\textbf{78.0}}           & \textbf{31.9}          & \textbf{25.0}          & \multicolumn{1}{c|}{\textbf{75.3}}           & \textbf{15.8}          & 34.2          & \multicolumn{1}{c|}{81.9}         & {21.4}          \\ \hline
\end{tabular}}
\vspace{-0.2cm}
\caption{Experiments on multilingual TATI task encompass translating Chinese into English, Korean, and Japanese.}
\vspace{-0.2cm}
\label{zh2all}
\end{table*}
\begin{table*}[t]
\resizebox{\textwidth}{!}{
\begin{tabular}{|c|ccc|ccc|ccc|}
\hline
\multirow{3}{*}{\textbf{Methods}}                                 & \multicolumn{3}{c|}{\textbf{en}$\rightarrow$\textbf{zh}}                                 & \multicolumn{3}{c|}{\textbf{ko}$\rightarrow$\textbf{zh}}                                 & \multicolumn{3}{c|}{\textbf{ja}$\rightarrow$\textbf{zh}}                                 \\ \cline{2-10} 
                                                                  & \multicolumn{2}{c|}{\textbf{I2T}}                   & \textbf{I2I}  & \multicolumn{2}{c|}{\textbf{I2T}}                   & \textbf{I2I}  & \multicolumn{2}{c|}{\textbf{I2T}}                   & \textbf{I2I}  \\ \cline{2-10} 
                                                                  & \textbf{BLEU} & \multicolumn{1}{c|}{\textbf{COMET}} & \textbf{BLEU} & \textbf{BLEU} & \multicolumn{1}{c|}{\textbf{COMET}} & \textbf{BLEU} & \textbf{BLEU} & \multicolumn{1}{c|}{\textbf{COMET}} & \textbf{BLEU} \\ \hline
nllb-200(3.3B)                                                    & 21.5          & \multicolumn{1}{c|}{73.3}           & 15.1          & 9.1           & \multicolumn{1}{c|}{65.3}           & 8.7           & 7.4           & \multicolumn{1}{c|}{61.3}           & 7.2           \\ \hline
m2m100(1.2B)                                                      & 24.2          & \multicolumn{1}{c|}{76.9}           & 18.9          & 14.8          & \multicolumn{1}{c|}{67.8}           & 13.1          & 24.3          & \multicolumn{1}{c|}{74.5}           & 22.7          \\ \hline
\begin{tabular}[c]{@{}c@{}}qwen1.5-7B-chat\end{tabular} & 27.6          & \multicolumn{1}{c|}{80.7}           & 21.4         & 20.9          & \multicolumn{1}{c|}{75.72}           & 18.2          & 30.0          & \multicolumn{1}{c|}{78.7}           & 27.5          \\ \hline
\begin{tabular}[c]{@{}c@{}}qwen1.5-14B-chat\end{tabular} & 34.5          & \multicolumn{1}{c|}{81.3}           & 26.8        & 27.7          & \multicolumn{1}{c|}{77.8}           & 23.6         & 38.4         & \multicolumn{1}{c|}{81.3}           & 28.6          \\ \hline
\begin{tabular}[c]{@{}c@{}}qwen1.5-110B-chat\end{tabular} & \textbf{37.9}          & \multicolumn{1}{c|}{84.2}           & 27.0          & 32.6          & \multicolumn{1}{c|}{80.5}           & 31.4          & 38.2          & \multicolumn{1}{c|}{80.7}           & 30.9          \\ \hline
\begin{tabular}[c]{@{}c@{}}qwen-max\end{tabular}          & 34.7          & \multicolumn{1}{c|}{84.1}           & 24.1          & 33.1          & \multicolumn{1}{c|}{81.0}           & 29.8          & 32.2          & \multicolumn{1}{c|}{80.4}           & 27.1          \\ \hline
\begin{tabular}[c]{@{}c@{}}qwen-vl-max\end{tabular}      & 36.3          & \multicolumn{1}{c|}{\textbf{84.3}}           & \textbf{27.8}          & \textbf{35.4}          & \multicolumn{1}{c|}{\textbf{81.7}}           & \textbf{31.6}          & \textbf{54.2}          & \multicolumn{1}{c|}{\textbf{83.8}}           & \textbf{44.3}          \\ \hline
\end{tabular}}
\vspace{-0.2cm}
\caption{Experiments on multilingual TATI tasks encompass translating English, Korean, and Japanese into Chinese.}
\vspace{-0.5cm}
\label{all2zh}
\end{table*}

\section{Experiments}

\subsection{Dataset}
\begin{figure}[h]
    \centering
    \includegraphics[width=\columnwidth]{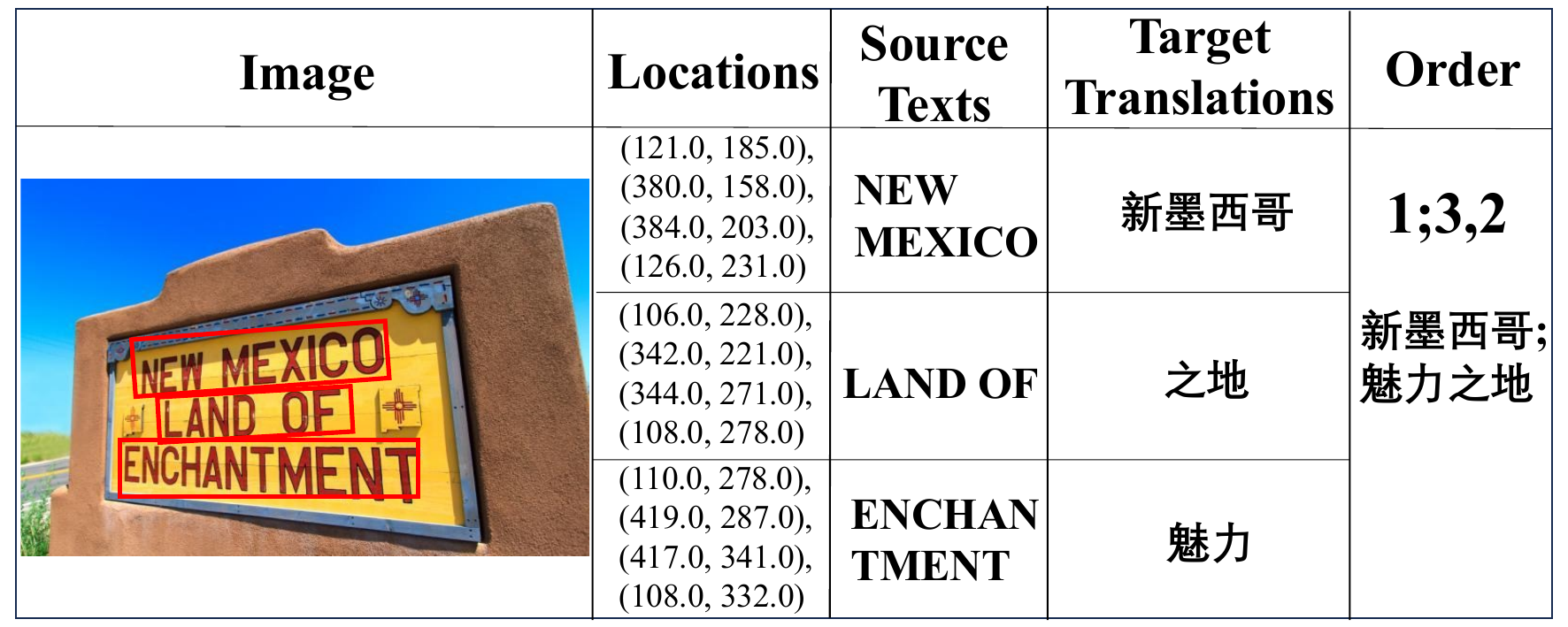}
    \vspace{-0.7cm}
    \caption{
An example of our MTIT6 dataset, which contains position information of the text in the image, corresponding translation information, and corrected translation order.}
    \vspace{-0.6cm}
    \label{fig:dataset}
\end{figure}
We present MTIT6, a comprehensive multilingual text image translation test dataset, assembled from ICAR 19-MLT\cite{Nayef_Liu_Ogier_Patel_Busta_Nath_Dimosthenis_Khlif_Matas_Pal_etal._2019}, OCRMT30K\cite{Lan_Yu_Li_Zhang_Luan_Wang_Huang_Su_2023}, along with a selection of high-quality images curated by our team. Our dataset encompasses six language pairs: English-to-Chinese, Japanese-to-Chinese, Korean-to-Chinese, Chinese-to-English, Chinese-to-Japanese and Chinese-to-Korean, each pair features about 200 images.
In creating this dataset, we employed the lightweight PP-OCR tool for initial OCR recognition, and then the OCR outputs were further refined and translated by language experts. 
Furthermore, considering  differences in word order  across different languages, our language experts meticulously annotated the sequences of text identified by OCR within each image. This approach enabled us to maintain semantic integrity by rearranging the text into coherent sequences, based on their annotated order. Figure~\ref{fig:dataset} presents an example of our MTIT6 dataset.


\subsection{Comparison Results}
\subsubsection{Quantitative Results}
For evaluation, we choose the \textbf{BLEU} \cite{Papineni_Roukos_Ward_Zhu_2001} and \textbf{COMET} \cite{Rei_Stewart_Farinha_Lavie_2020} metrics. We evaluate the image-to-text (\textbf{I2T}) intermediate translation results and image-to-image (\textbf{I2I}) final translation results. We have integrated a wide range of models into our \modelname, which included classic encoder-decoder models \cite{costa2022no,fan2021beyond}, widely accessible open-source LLMs (qwen-chat1.5-7B,14B and 110B), and commercially advanced close-source LLM (qwen-max) and VLM \cite{bai2023qwen} (qwen-vl-max), affirming our approach's comprehensive reliability and easy scalability. We also validate the model\cite{Lan_Yu_Li_Zhang_Luan_Wang_Huang_Su_2023} specifically designed for the TIT task in our test dataset. To more accurately evaluate the translation quality of the final image, we use the paid BaiduOCR\footnote{\scriptsize{\url{https://cloud.baidu.com/product/ocr/general}}} to recognize the text in the \textbf{I2I} stage. 

As shown in Table~\ref{zh2all} and Table~\ref{all2zh}, we observed that the performance of the qwen-1.5 series models gradually improved with the increase of the model's parameters. We discovered that the enhancement in performance is attributed not only to the improved quality of translations but also to the bolstered ability to follow instructions. This is particularly evident in the 7B model, which initially exhibited a weaker capacity for instruction adherence. During qwen-7B model's translation process, there is around a 10\% chance that the \texttt{<box{idx}></box{idx}>} symbol, employed to demarcate positions, might be inaccurately translated. Another interesting finding is that the performance of qwen1.5-110B is very close to or even exceeds qwen-max in multiple language pairs. This may be because the qwen1.5 series used more new high-quality corpora and adopted technologies such as DPO\cite{Rafailov_Sharma_Mitchell_Ermon_Manning_Finn} and PPO\cite{Schulman_Wolski_Dhariwal_Radford_Klimov_2017} during training. The results demonstrate that while enlarging the model's parameters significantly boosts its capability to adhere to instructions, honing the model's translation skills may rely more heavily on the quality of the corpus and the refinement of training methodologies.
Moreover, VLMs improved translation performance, indicating that integrating image data can further augment translation accuracy. This advancement confirms that VLMs represent a key developmental trajectory for future research endeavours in image translation.

\begin{figure}[t]
    \centering
    \includegraphics[width=\columnwidth]{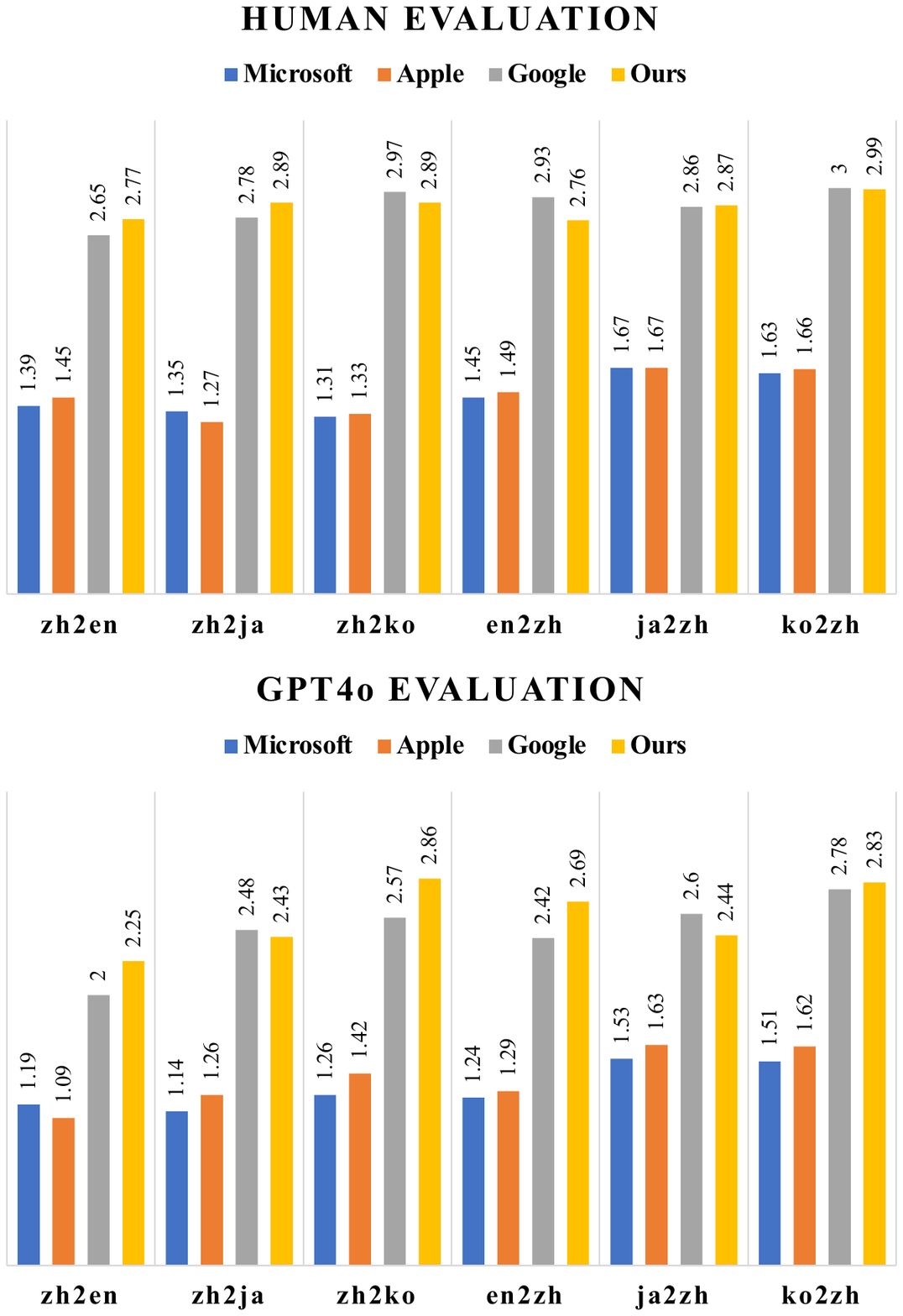}
    \vspace{-0.6cm}
    \caption{
Overall human evaluation and GPT4o results of image translation performance for different methods. Our method significantly outperforms Microsoft and Apple and achieves comparable results to Google.}
    \vspace{-0.3cm}
    \label{fig:human_gpt4o}
\end{figure}

\begin{figure}[]
    \centering
    \includegraphics[width=\columnwidth]{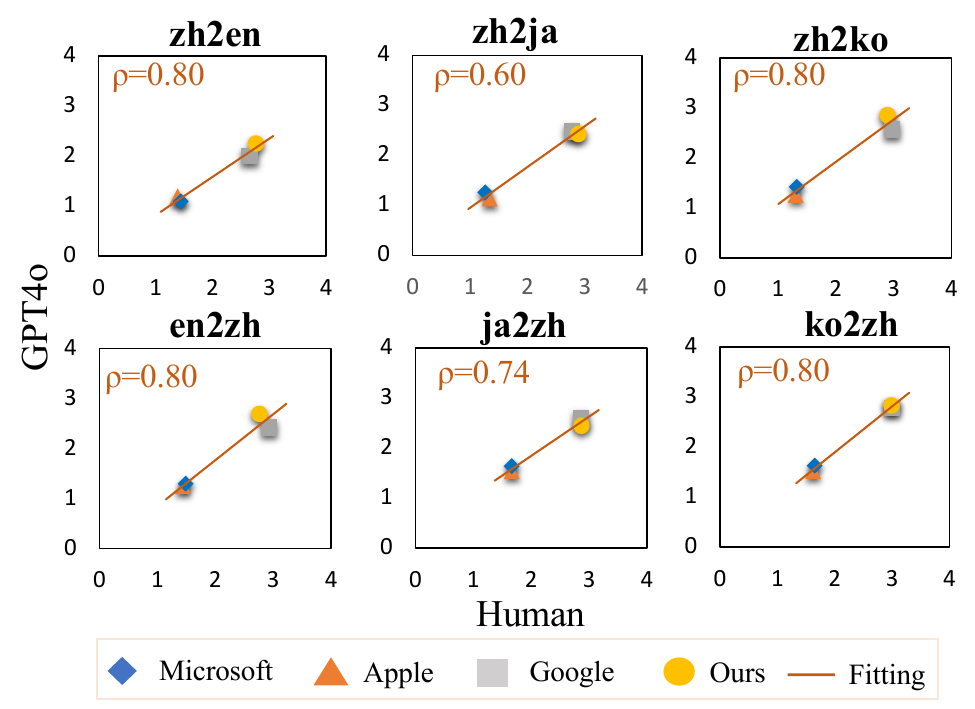}
    \caption{
Our experiments show that GPT4o evaluations across all language pairs closely match human perceptions. In each plot, a dot represents the
human preference evaluation score (horizontal axis) and GPT4o evaluation score (vertical axis). We
linearly fit a straight line to visualize the correlation and calculate Spearman’s correlation coefficient ($ \rho $) for each language pair.}
    \label{fig:Spearman}
\end{figure}

\subsubsection{Qualitative Results}
\begin{figure*}[!h] 
  \centering 
  \includegraphics[width=2.1\columnwidth]{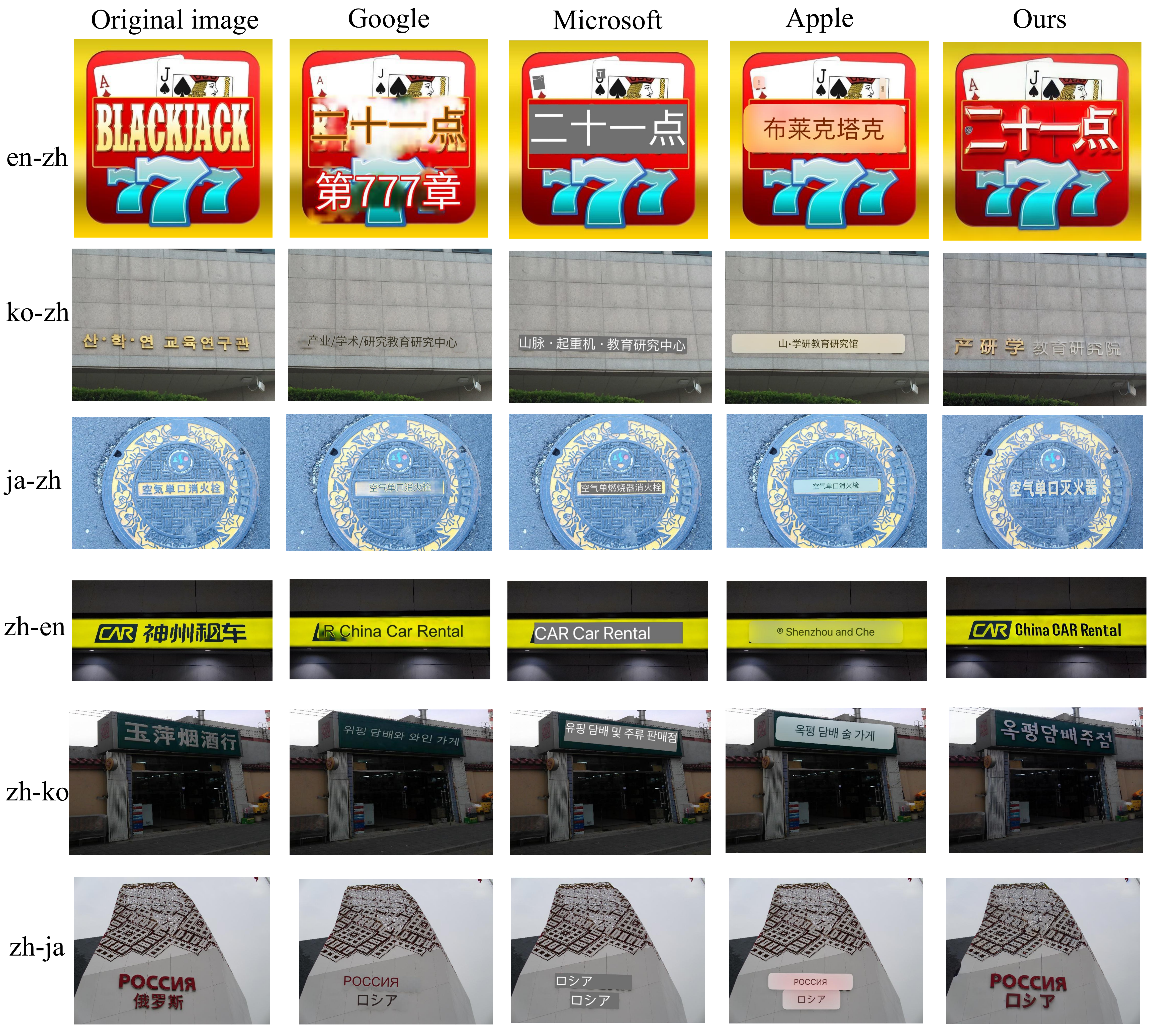} 
  \vspace{-0.5cm}
  \caption{Qualitative comparison of our framework with Google, Microsoft and Apple Image Translation results. Our \modelname has obvious advantages in font style preservation and authenticity.}
  \vspace{-0.5cm}
  \label{fig:visualize1}
\end{figure*}

To the best of our knowledge, this is the first paper to research the task of \taskname, so there is no open-source model to compare with, so we can only compare with commercial closed-source image translation products, including Google Image Translation, Microsoft Image Translation and Apple IOS Image Translation. As shown in the cases in Figure~\ref{fig:visualize1}, Microsoft and Apple Image Translation generate translations in rectangular areas based on rules and then paste them back to the original image. However, these rectangular areas' colours fail to match those of the original image. Consequently, directly integrating the text from these areas into the original image significantly disrupts its visual harmony. Google Image Translation exhibits some improvement. It first erases the original text and then returns the translated text to the original image. However, this process leaves noticeable erasure marks, and the text, being rule-based, appears overly uniform and fails to harmonize with the original image's aesthetics.
In contrast, our \modelname seamlessly integrates the translated text into the original image and even manages to preserve the font colour and style to a notable degree. Therefore, it is clear that our \modelname significantly surpasses image translation products in maintaining visual continuity.

\subsubsection{Human and GPT Evaluation}
To evaluate the authenticity and style consistency of translated images, we randomly selected 50 images from six language pairs, totalling 300 images. We then assessed the translation results from Google Image Translation, Microsoft Image Translation, Apple Image Translation, and \modelname. Each image was scored based on our evaluation criteria by three assessors and GPT4o, and the detailed evaluation criteria can be found in the appendix. As shown in Figure~\ref{fig:human_gpt4o}, whether it is the human evaluation or GPT4o automatic evaluation, our method significantly outperforms Microsoft and Apple Image Translation in terms of authenticity and style consistency and achieves comparable scores to Google. 
We also verify the correlation between GPT4o evaluation results and human preference scores in Figure~\ref{fig:Spearman}. By calculating Spearman’s correlation coefficient for each language pair, we observe a strong correlation between the two evaluation methods. The consistency further demonstrates the superiority of our approach.

Upon analyzing the cases with lower scores than Google, we found most instances are due to the limited performance of \modelname in generating text on small fonts. In contrast, Google Image Translation, being based on rule-based generation of text, has a clear advantage in translating texts of small font sizes. Nevertheless, based on the advantages of authenticity and style consistency, our \modelname still achieved scores comparable to Google Image Translation.

\begin{table}[t]
\resizebox{0.48\textwidth}{!}{
\begin{tabular}{|c|cc|}
\hline
\textbf{Methods}                                                           & \multicolumn{2}{c|}{\textbf{Average}} \\ \cline{2-3} 
                                                                           & \textbf{BLEU}     & \textbf{COMET}    \\ \hline
\begin{tabular}[c]{@{}c@{}}qwen1.5-7B-chat(box)\end{tabular}   & 25.9              & 75.7              \\ \hline
\begin{tabular}[c]{@{}c@{}}qwen1.5-7B-chat(context)\end{tabular}    & 26.5              & 76.3             \\ \hline
\begin{tabular}[c]{@{}c@{}}qwen1.5-14B-chat(box)\end{tabular}   & 30.6              & 76.9              \\ \hline
\begin{tabular}[c]{@{}c@{}}qwen1.5-14B-chat(context)\end{tabular}    & 31.0              & 77.9             \\ \hline
\begin{tabular}[c]{@{}c@{}}qwen1.5-110B-chat(box)\end{tabular} & 32.2              & 78.1             \\ \hline
\begin{tabular}[c]{@{}c@{}}qwen1.5-110B-chat(context)\end{tabular}    & \textbf{33.2}   & \textbf{79.1}            \\ \hline
\end{tabular}}
\vspace{-0.2cm}
\caption{Ablation experiments on translation strategies and model categories on multilingual TIT tasks.}
\vspace{-0.6cm}
\label{ablation_boxbybox}
\end{table}
\begin{table}[]
\resizebox{0.48\textwidth}{!}{
\begin{tabular}{|c|ccc|}
\hline
\multirow{3}{*}{\textbf{Methods}}                                         & \multicolumn{3}{c|}{\textbf{zh}$\rightarrow$\textbf{en}}                                 \\ \cline{2-4} 
                                                                          & \multicolumn{2}{c|}{\textbf{I2T}}                   & \textbf{I2I}  \\ \cline{2-4} 
                                                                          & \textbf{BLEU} & \multicolumn{1}{c|}{\textbf{COMET}} & \textbf{BLEU} \\ \hline
\begin{tabular}[c]{@{}c@{}}qwen1.5-110B-chat\end{tabular}           & 43.8          & \multicolumn{1}{c|}{76.27}          & \textbf{30.6}          \\ \hline
\begin{tabular}[c]{@{}c@{}}Wo-resize\end{tabular} & 43.8          & \multicolumn{1}{c|}{76.27}          & 27.7(\textbf{-2.9})    \\ \hline
\end{tabular}}
\caption{Ablation experiment on resizing editing area.}
\vspace{-0.7cm}
\label{ablation_woresize}
\end{table}
\subsection{Ablation Study}
We performed detailed ablation studies to explore the efficacy of two translation strategies: translating the contents within detection boxes individually versus translating all recognized text in the image as a whole. Specifically, for the latter translation method, we concatenate recognized texts from an entire image using `<box{idx}></box{idx}>` tags. These are then merged with few-shot prompts into a lengthy sentence, which is subsequently inputted into LLMs for translation. We tested on the qwen1.5-7B, 14B and 110B models and calculated the average of the test results for all language pairs. As depicted in Table~\ref{ablation_boxbybox}, our strategy of translation as a whole significantly improves translation performance across all three parameter sizes of qwen1.5 models. This enhancement underscores the importance of LLM's advanced contextual understanding in boosting translation performance.
We also conducted an ablation experiment on the resize editing area strategy. As shown in Table~\ref{ablation_woresize}, in the zh2en translation, without the OCR box resizing 
step, the final I2I translation result dropped by 2.9 points, proving the effectiveness of this strategy.

\section{Discussions}
As the first paper to introduce (vision) LLMs and diffusion model into the Translate AnyText in the Image (\taskname) task, significant opportunities exist for further improvement. Below, we enumerate several potential directions for future advancements:

\paragraph{(1)}Integration of OCR and Translation Processes: Our current methodology bifurcates the process into OCR text recognition and translation as distinct steps. While VLMs currently fail to achieve the OCR accuracy of smaller models tailor-made for OCR tasks, further development and OCR-targeted training could potentially elevate VLMs to achieve formidable OCR prowess. This evolution could potentially consolidate text recognition and translation into a seamless, singular step, enhancing efficiency and accuracy.
\paragraph{(2)}Text editing model adapted to translation: Due to AnyText~\cite{tuo2023anytext} being trained on datasets where character size perfectly matches the image size, it needs the text length to be well-matched with the dimensions of the editing area. However, when translating, the length of the translated text inevitably varies across different languages, leading to challenges for Anytext to generate translations that fit the original text area perfectly. The Anticipated Box Resizment strategy helps mitigate the issue but does not fully resolve it. Future efforts could focus on training a text editing model capable of dynamically adjusting font sizes. This would eliminate the necessity for altering the editing area, allowing for modifications that preserve the aesthetic appeal and structural harmony of the original image more faithfully.
\section{Conclusion}
We introduce a novel framework named \modelname designed for Translate AnyText in the Image (\taskname). Distinguished from existing closed-source products, our \modelname can be built upon open-source models and is training-free. Uniquely, we integrate (vision) LLMs and diffusion models into \taskname task for the first time, achieving both accurate translations and authentic translated images. Furthermore, we have curated a multilingual text image translation dataset MTIT6 to promote development in this field.

\newpage
\section{Limitations}
\paragraph{(1)}
Owing to inherent restrictions in Anytext~\cite{tuo2023anytext}, it is unable to produce outputs exceeding 20 letters or characters at a time. Consequently, this limitation extends to our AnyTrans, affecting its ability to effectively translate longer texts.
\paragraph{(2)}
Given that Anytext's text editing proficiency is confined to Chinese, English, Korean, and Japanese, it lacks the capability to generate text in other languages, such as Arabic. As a result, the range of languages that \modelname is capable of translating is similarly restricted.
\bibliography{custom}

\newpage
\appendix
\section{Appendix}
\subsection{Dataset annotation details}
We engaged six professional translators for a week-long annotation task, with each translator tasked to annotate 30 images daily to mitigate fatigue. Upon being presented with an image containing source texts detected by PPOCR-v4, translators were tasked to render accurate and fluid translations into the target language. Furthermore, they meticulously annotated the sequences of text recognized by OCR within each image, reordering the text to ensure coherent sequences. For quality assurance, we also employed a professional translator to sample and review the annotated instances. In total, we annotated 1,199 images, averaging around 200 instances per language pair.
\subsection{Human and GPT evaluation details}
We meticulously selected a sample of 50 images for each of the six languages, summing up to a total of 300 images. To objectively and accurately assess the \textbf{authenticity} of translated images along with the \textbf{maintenance of font styles}, we utilize a combination of human evaluation and GPT-4o evaluation.

For human evaluation, we enlisted the help of three annotators. For each image assessed, the annotators were provided with the original image alongside the translation outputs from Google, Microsoft, Apple Image Translations, and our AnyTrans. They then scored each translation based on predetermined criteria, with the final score for each image being the average of the three annotators' scores.

For the evaluation involving GPT-4o, to minimize biases associated with the order in which translations are presented, the evaluation is conducted on a one-to-one basis: compare the source image with the translated image from one of the four different methods. This approach was adopted to impartially assess the effectiveness of the four image translation methodologies.

For both human and GPT-4o powered evaluations, detailed results are provided in the supplementary materials, which include the specific images evaluated and the resulting scores. The detailed evaluation criteria are outlined as follows:
\paragraph{(1)}\textbf{1 point}
Very low authenticity: The translated text looks completely unnatural and clearly distinguished from the background of the image as if it was added randomly.
Inconsistent style: Ignoring the font, size, color and position of the original text, the inconsistency in style makes the entire translated image feel unreal or abrupt.
\paragraph{(2)}\textbf{2 points}
Low authenticity: The translated text is slightly stiff in the image and lacks a sense of integration. It can be clearly seen that it was added later.
Partially coordinated style: The translated text tries to imitate the original style to a certain extent, but the overall effect is not good, and the sense of style is more obvious.
\paragraph{(3)}\textbf{3 points}
General authenticity: The translated text is relatively natural and can be integrated into the image to a certain extent, but there are still recognizable inconsistencies.
Partially coordinated style: The translated text partially echoes the style of the original image and contains the correct elements (such as font, size, color), but still lacks some overall harmony.
\paragraph{(4)}\textbf{4 points}
High authenticity: The translated text is well integrated into the image, giving people a more natural feeling, and only small flaws may be found when looking closely.
Generally coordinated style: The style of the text matches the original image to a large extent. Small details can be optimized, but the overall look and feel is close to the same.
\paragraph{(5)}\textbf{5 points}
High authenticity: The translated text blends perfectly with the image background, and it is almost impossible to tell that the text was added later.
Completely coordinated style: The style is completely consistent with the original text, including font, size, color, position and shadow effects, and the overall effect is coordinated and very professional.

In actual evaluation, these two aspects can be considered comprehensively based on the overall effect of the translated image on the score.

\end{document}